\pdfoutput=1

\documentclass[11pt]{article}

\usepackage{acl2023}

\usepackage{times}
\usepackage{latexsym}

\usepackage[T1]{fontenc}
\usepackage{xcolor,colortbl}
\usepackage{tabu}
\usepackage{makecell}

\usepackage[utf8]{inputenc}

\usepackage{microtype}

\usepackage{inconsolata}

\usepackage{graphicx}
\usepackage{subfigure}
\usepackage{booktabs} 
\usepackage{amsmath}
\usepackage{amssymb}
\usepackage{amsfonts}

\newcommand{\Dc}{\mathcal{D}}

\newcommand{\xv}{\mathbf{x}}
\newcommand{\yv}{\mathbf{y}}

\usepackage[disable]{todonotes} 

\title{Learning Mutually Informed Representations for Characters and Subwords
}

\author{Yilin Wang$^*$ \\
  Harvard University \\
  \texttt{yilin\_wang@g.harvard.edu} \\\And
  Xinyi Hu$^*$ \\
  Carnegie Mellon University \\
  \texttt{xinyih2@alumni.cmu.edu} \\ \And
  Matthew Gormley \\
  Carnegie Mellon University \\
  \texttt{mgormley@cs.cmu.edu} \\ 
  }

\begin{document}
\maketitle

\newcommand{\wordside}{\textsc{Subw}}
\newcommand{\charside}{\textsc{Char}}

\begin{abstract}
Most pretrained language models rely on subword tokenization, which processes text as a sequence of subword tokens. However, different granularities of text, such as characters, subwords, and words, can contain different kinds of information. Previous studies have shown that incorporating multiple \emph{input} granularities improves model generalization, yet very few of them \emph{outputs} useful representations for each granularity. In this paper, we introduce the \emph{entanglement model}, aiming to combine character and subword language models. Inspired by vision-language models, our model treats characters and subwords as separate modalities, and it generates mutually informed representations for \emph{both} granularities as output. We evaluate our model on text classification, named entity recognition,  POS-tagging, and character-level sequence labeling (intraword code-switching). Notably, the entanglement model outperforms its backbone language models, particularly in the presence of noisy texts and low-resource languages. Furthermore, the entanglement model even outperforms larger pre-trained models on all English sequence labeling tasks and classification tasks. We make our code publically available.\footnote{\url{https://github.com/TonyW42/noisy-IE}}
\end{abstract}

\section{Introduction}






Since the emergence of pretrained language models (LMs) like ELMo \cite{peters-etal-2018-deep} and BERT \cite{devlin-etal-2019-bert}, subwords tokenization have become the prevailing approach to tokenization. Common techniques include byte-pair-encoding (BPE) \cite{sennrich-etal-2016-neural}, WordPiece \cite{wu2016googles}, and SentencePiece \cite{kudo-richardson-2018-sentencepiece}, which create word-sized character n-grams for the LM to learn reusable representations. However, subword tokenization has limitations: the number and vocabulary of subwords must be predetermined during pretraining. Consequently, tasks involving noisy text or low-resource languages often require meticulous engineering to achieve satisfactory performance.

A less studied alternative is tokenizing at the character or byte level. Pretrained LMs like CANINE \cite{clark-etal-2022-canine}, Charformer \cite{tay2022charformer}, and ByT5 \cite{xue-etal-2022-byt5} utilize character-level tokenization. Though such models usually require careful design to handle longer sequences resulting from fine-grained tokenization, they offer advantages such as better incorporation of morphology and avoidance of tokenization overfitting to the pretraining corpus domain.

Previous studies have shown that incorporating both character and subword (or full word) representations can enhance model generalization. However, most studies focused on using characters to enhance or refine word representations \cite{aguilar-etal-2018-modeling,sanh_hierarchical_2019,shahzad-etal-2021-inferner,wang-etal-2021-automated, ma-etal-2020-charbert, tay2022charformer}. However, these models, unlike the character-level pretrained language models mentioned earlier, do not generate usable character-level representations.

In this paper, we argue that character and subword representations are distinct yet complementary. We introduce a novel model, named the \emph{entanglement model}, which combines a pretrained character LM and a pretrained subword LM. Inspired by techniques from the vision-language models (specifically ViLBERT \cite{lu2019vilbert}), we treat characters and subwords as two modalities and leverage cross-attention to learn new representations by iteratively attending between the character and subword sides of the model.  The result is a simple, yet general approach for bringing together the fine-grained representation afforded by characters with the rich memory of subword representations. 

We evaluate our entanglement model on a variety of \emph{tasks} (named entity recognition (NER), part-of-speech (POS) tagging, and sentence classification), \emph{domains} (noisy and formal text), and \emph{languages} (English and ten African languages). We also evaluate the entanglement model on character-level tasks (intraword code-switching), which cannot be processed by subword models.
Empirically, our model consistently outperforms its backbone models and previous models that incorporate character information. On English sequence labeling and classification tasks, the entanglement model even outperforms larger pre-trained models. Further, we found that the usage of subword-aware character representations yields performance gains, compared to using a character-only model. 

%
%

In order to better understand the effectiveness of our model, we also explore two natural extensions: (1) incorporating positional embeddings that explicitly align the characters and subwords and (2) masked language model (MLM) pretraining of the entanglement model. We find that these augmentations of the model are unnecessary, suggesting that our model is capable of learning positional alignment between characters and subwords on its own and leveraging the substantial pretraining of the backbone models \emph{without} costly pretraining of our entanglement cross-attention layers.

\section{Methods}

\begin{figure*}[h]
    \centering
    \includegraphics[width = 14cm]{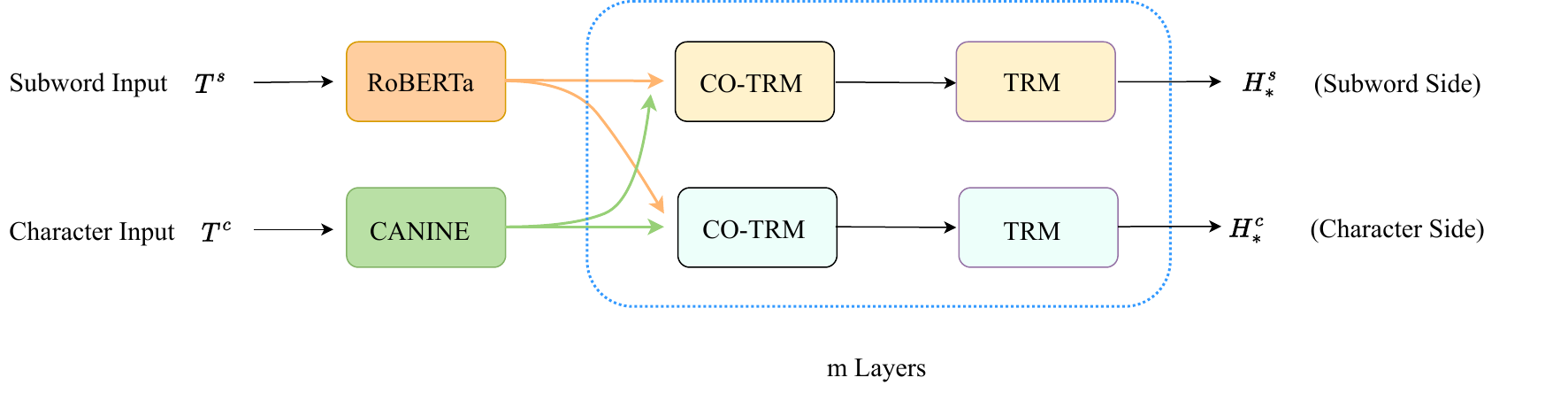}
    \caption{Architecture of the entanglement model.}
    \label{fig:alt}
\end{figure*}

We propose a novel \emph{entanglement model} that allows information exchange between pretrained character models and subword models, which is facilitated by two separate sets of co-attention modules. Our intention is for each layer of co-attention to further entangle the subword and character representations. 
The model thereby builds subword representations that are character-aware and character representations that are subword-aware
which can be used on both character-level and word-level tasks. 

We apply the model to sequence labeling and text classification 
assuming a dataset of $N$ samples and $K$ classes, 
$\Dc = \{(\xv^{(i)}, y^{(i)})\}_{i=1}^N$,
where $\xv^{(i)} \in \mathbb{R}^{n_i}$ is a sequence of words of length $n_i$ with label $\yv^{(i)}$. For sequence labeling, the label $\yv^{(i)} \in \{1, 2, \cdots, K\}^{n_i}$ is a vector with the same length as $\xv^{(i)}$. For text classification, the label $\yv^{(i)} \in \{1, 2, \cdots, K\}$ is an integer. 

\subsection{The Entanglement Model}

Figure \ref{fig:alt} shows the architecture of the entanglement model.
We describe the model for a single training example $(\xv, \yv)$, we first tokenize it into a subword sequence $\xv^{s} \in \mathbb{R}^{n^s}$ and a character sequence $\xv^{c} \in \mathbb{R}^{n^c}$, where $n^s, n^c$ refers to the length of the subword and character sequences respectively. We then feed $\xv^{c}$ through a character encoder and $\xv^{s}$ through a subword encoder to obtain contextualized representations $H^{s} \in \mathbb{R}^{n^s \times d}$ and $H^{c} \in \mathbb {R}^{n^c \times d}$, where $d$ is the embedding size for the contextualized representations. Then, we feed $H^{s}$ and $H^{c}$ through $m$ (separate) co-attention modules to facilitate information exchange between character and subword representations, which outputs a character-aware subword embedding $H^s_*$ and a subword-aware character embedding $H^c_*$. 
When using $H^{s}_*$ for inference, we call the experiment to use the \emph{subword side} ($\wordside$). When using $H^{c}_*$ for inference, we call the experiment to use the \emph{character side} ($\charside$)

While having separate encoders for characters and subwords allows better modeling of the features unique to each granularity, the cross-attention block inside the co-attention module allows the representations for characters and words to learn from each other. During training, the information exchange happens not only in the co-attention modules but also in the backbone text encoders through the flow of the gradient. 


\subsection{The Co-attention Module}
\begin{figure}[t]
    \centering
    \includegraphics[width = 6.3cm]{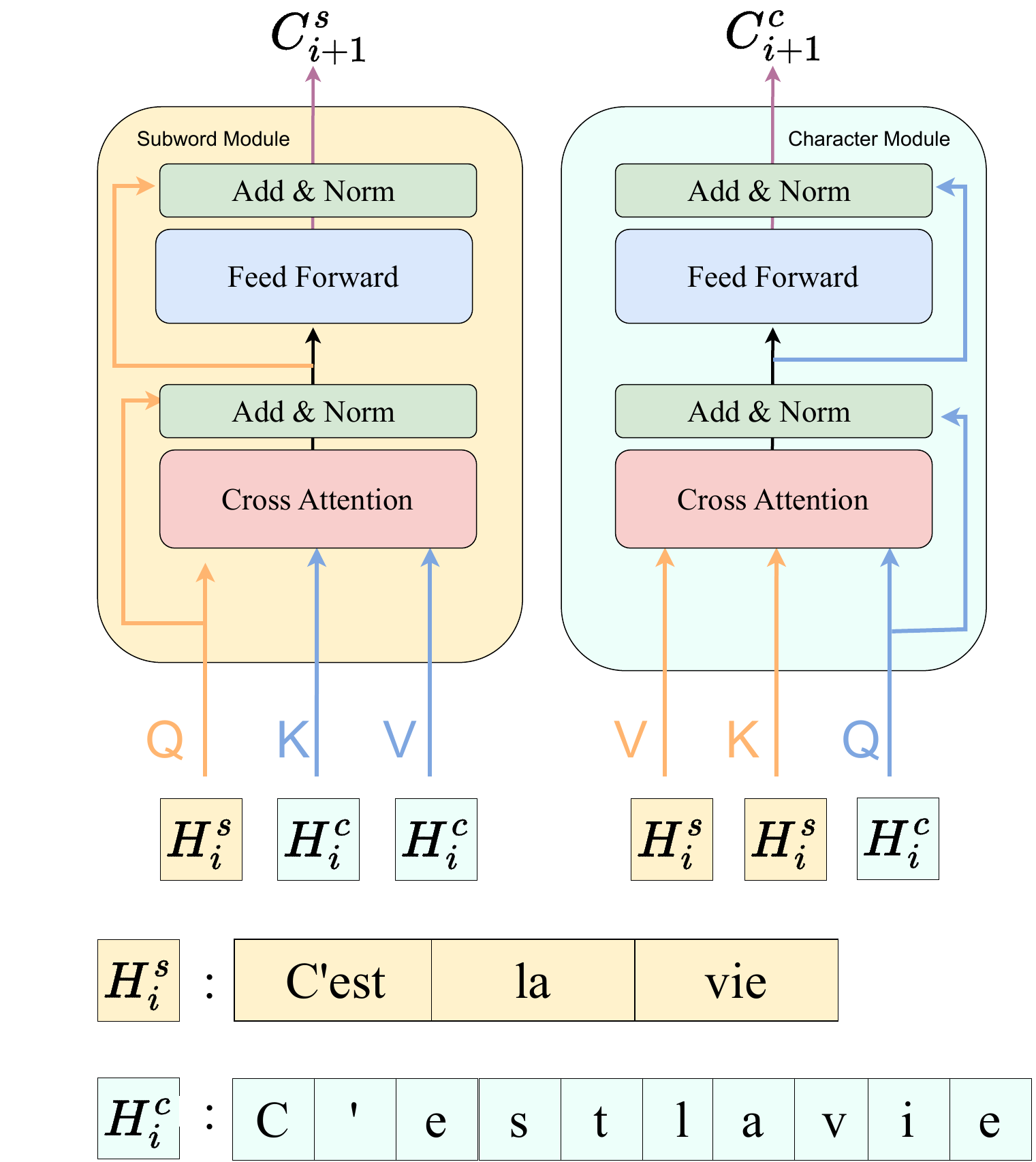}
    \caption{Architecture of the CO-TRM block inside the co-attention module.}
    \label{fig:cotrm}
\end{figure}
A co-attention module consists of two transformer blocks \cite{https://doi.org/10.48550/arxiv.1706.03762}. The first transformer block, named CO-TRM, features a cross-attention layer that uses one modality to query the other, which facilitates information exchange between the two modalities. Figure \ref{fig:cotrm} demonstrates the structure of the CO-TRM module. The second transformer block, named TRM, features a self-attention layer, which is the same as the transformer layers in the backbone encoders.

Let $H^s_0 = H^s$ and $H^c_0 = H^c$ be the output of the pretrained LMs and $H^s_i$ and $H^c_i$ be the subword and character embeddings output by the $i^{th}$ co-attention module. Given $H^s_i$ and $H^c_i$ the subword-side co-attention module outputs the next-layer hidden states $H^s_{i+1}$ as:
\begin{align*}
C^s_{i+1} &= \text{CO-TRM}(Q = H^s_i, K = H^c_i, V = H^c_i) \\ 
H^s_{i+1} &= \text{TRM}(Q = K=  V = C^s_{i+1})
\end{align*}
Where $C^s_{i+1}$ refers to the intermediate representation output by the CO-TRM module. Similarly, the character side co-attention module outputs the next-layer hidden states $H^s_{i+1}$ as:
\begin{align*}
C^c_{i+1} &= \text{CO-TRM}(Q = H^c_i, K = H^s_i, V = H^s_i) \\ 
H^c_{i+1} &= \text{TRM}(Q =K = v = C^c_{i+1})
\end{align*}

\subsection{Sequence Labeling}
\label{sec:sequence-labeling}


\paragraph{The Subword Side}
When training our model through the subword side, we pass the character-aware subword embedding $H^s_*$ through a linear classification layer and a softmax layer to obtain the output probabilities $\hat{p}^s \in \mathbb{R}^{n_s \times K}$ for each subword:
$$
\hat{p}^s = \text{Softmax}\left( H^s_* W^s\right) \quad \quad W^s \in \mathbb{R}^{d \times K}
$$
We then select the output probabilities for the first subword as the prediction for each word, which creates word-level output probabilities $\hat{p}^{w} \in \mathbb{R}^{n \times K}$.

\paragraph{The Character Side}
Similarly, when training our model on the character side, we use the subword-aware character embedding $H^c_*$ to obtain the output probabilities $\hat{p}^c \in \mathbb{R}^{n_c \times K}$ for each character:
$$
\hat{p}^c = \text{Softmax}\left( H^c_* W^c\right) \quad \quad W^c \in \mathbb{R}^{d \times K}
$$
We then select the output probabilities for the first character as the prediction for each word to get word-level output probabilities $\hat{p}^{w} \in \mathbb{R}^{n \times K}$. 

\paragraph{Loss and Inference}
We then train the model under cross-entropy loss: 
$$
\mathcal{L}(\hat{p}^{w}, \boldsymbol{y}) = \sum_{j=1}^{n} \sum_{k = 1}^K \boldsymbol{y}^{j}_k \log(\hat{p}^{w, j}_k)
$$
where $\boldsymbol{y}^{j}_k$ refers to the one-hot encoding of the label on the word $j$ and $\hat{p}^{w, j}_k$ refers to output probability of that word on class $k$. 

For inference, we will take the class with the highest output probability as the predicted label for each word. i.e., 
$$
\hat{y}^{j} = \text{argmax}_k \quad \hat{p}^{w, j}_k 
$$

\subsection{Text Classification}
\label{sec:text-classification}

\paragraph{The Subword and Character Sides}

For text classification, the procedure for the subword side and the character side is the same: We take $h \in \mathbb{R}^d$, the first argument of either $H^s_*$ or $H^c_*$, which is the embedding for the \verb+[CLS]+ token, and pass it through a linear and $\tanh$ layer. We then pass this output through a linear classification layer and a softmax function to obtain the output probabilities $\hat{p} \in \mathbb{R}^K$:
$$
\hat{p} = \text{Softmax}(W^c(\sigma (W^p h)))
$$
where $W^p \in \mathbb{R}^{d \times d}$, $W^c \in \mathbb{R}^{d \times K}$, and $\sigma(\cdot)$ refers to the $\tanh(\cdot)$ function. 

\paragraph{Loss and Inference}

We then train the model under cross-entropy loss: 
$$
\mathcal{L}(\hat{p}, \boldsymbol{y}) =  \sum_{k = 1}^K \boldsymbol{y} \log(\hat{p}_k)
$$
where $\boldsymbol{y}$ refers to the one-hot encoding of the label of the sample text and $\hat{p}_k$ refers to output probability of that sample on class $k$. 

For inference, we take the class with the highest output probability as the predicted label for each sample. i.e., 
$$
\hat{y} = \text{argmax}_k \quad \hat{p}_k 
$$

\subsection{Comparison with Previous Work}
Our model architecture draws partial inspiration from ViLBERT \citep{NEURIPS2019_c74d97b0}, a pretrained vision-language model. However, unlike ViLBERT, our model capitalizes on the capabilities of pretrained character and subword models, eliminating the need for additional pretraining steps and resulting in faster training times.

Other studies have investigated combining character and word embeddings. The ACE model \cite{wang-etal-2021-automated} uses neural architecture search to find a subset of 11 embeddings,  which are concatenated to form word representations. Unlike our model, ACE relies on fixed word embeddings, lacks learned character representations, and requires computationally intensive search. Our model is more efficient and learns a fine-grained representation of characters and subwords.

\section{Experimental Setup}
\subsection{Datasets and Tasks}
We evaluate our model on four tasks: named entity recognition (NER), part-Of-Speech (POS) tagging, intraword code-switching and text classification.

\textbf{English sequence labeling}: For NER, We utilize the WNUT-17 dataset \citep{derczynski-etal-2017-results} and the CONLL-2003 dataset \citep{tjong-kim-sang-de-meulder-2003-introduction}, which respectively contains noisy user-generated texts from social media and formal writings sourced from the Reuters news. For POS-tagging, we use TweeBank \citep{jiang-etal-2022-annotating}, which contains noisy texts from Twitter.


\textbf{Multilingual NER}: We use the MasakhaNER dataset \citep{adelani-etal-2021-masakhaner}, which offers NER tasks for 10 low-resourced African languages. 

\textbf{Character-level sequence labeling:}
We also use the Spanish-Wixarika and Turkish-German data of \citet{mager2019subwordlevel} on \emph{intraword} code-switching. Since the language switch exists within a word, the \emph{intraword} segmentations cannot be predicted by subword models because the morpheme boundaries might not align with subword boundaries. We formulate it as a character-level sequence labeling task. 


\textbf{Text classification}: The WNUT-2020 shared task \#2 dataset \citep{nguyen2020wnut2020} focuses on identifying informative English tweets related to COVID-19. Additionally, we use the TweetEval dataset \citep{barbieri-etal-2020-tweeteval}, a comprehensive benchmark for evaluating tweet classification.






\subsection{Experimental Details}
For our experiments, we utilize the CANINE-s\footnote{The best CANINE model from \cite{clark-etal-2022-canine} employs character n-gram embeddings. However, the corresponding pretrained model is not released by Google, so we use the available model: CANINE-s.} \citep{clark-etal-2022-canine} as the underlying character encoder backbone. For multilingual sequence labeling tasks, we employ $\text{XLM-R}_{\text{base}}$ \citep{conneau-etal-2020-unsupervised} as the subword encoder backbone, while for all other tasks, we use $\text{RoBERTa}_{\text{base}}$ \citep{zhuang-etal-2021-robustly} as the subword encoder backbone. 

During model training, we employ the Adam optimizer with an initial learning rate of 2e-5 and a linear scheduler. The number of maximum epochs varies for each dataset: 25 for TweetEval and 50 for all other datasets. We select the model with the best performance on the validation set and evaluate it on the test set. Due to the small scale of MasakhaNER, we run each experiment three times with different seeds and report the average results.

We evaluate the entanglement model against four baselines: the backbone text and character model, a larger pre-trained subword model, and CharBERT \cite{ma-etal-2020-charbert}, a previous subword model that incorporates character information. 

%
In our result tables, we employ bold to highlight the best outcome achieved by either our baselines or the entanglement model, while $\dagger$ denotes the state-of-the-art performance. We keep the numbers from prior work in greyscale in all following tables.

\section{Results}



We conduct an extensive analysis of our model's performance on various sequence labeling and text classification tasks. We evaluate the effectiveness of our model on both formal and noisy English texts, as well as low-resourced languages, in order to assess its capabilities across different scenarios. Moreover, for each task, we report the performance of different configurations of our model, such as utilizing the subword or character side and varying the number of co-attention modules. This approach enables us to examine the robustness of our modules under different hyperparameter settings.

\subsection{English Sequence Labeling}


\begin{table}[t]
    \centering
    \begin{tabu}{ l  c c  c}
    \toprule
        \multicolumn{2}{l}{\textbf{Model}} &  \textbf{WNUT-17} & \textbf{CONLL-03} \\ \midrule
        \rowfont{\color{gray}}
        \multicolumn{2}{l}{ ACE 
        } & - & 94.60$^\dagger$\\
        \rowfont{\color{gray}}
        \multicolumn{2}{l}{CL-KL} & 60.45$^\dagger$ & -\\
        \multicolumn{2}{l}{$\text{RoBERTa}_{\text{large}}$} & 57.10 & \color{gray} 92.31 \\
        \multicolumn{2}{l}{$\text{RoBERTa}_{\text{base}}$} & 56.38 & 91.93\\
        \multicolumn{2}{l}{CharBERT} & 53.63 & 92.07 \\
        \multicolumn{2}{l}{CANINE-s} & 24.27 & 86.23 \\ \midrule\midrule 
        \textbf{Side} & \textbf{\#C} & & \\ \midrule
        \charside & 1 & 40.45 & 89.09\\
        & 2  & 39.77 & 89.57 \\
        &3  & 39.46 & 89.43 \\
         & 4  & \textbf{42.42} & \textbf{89.74}\\ 
         \midrule
        \wordside & 1  & 57.80 & 91.81\\
        & 2  & \textbf{57.97} & 92.21\\
        &3  & 57.14 & 92.07\\
         & 4  & 56.28 & \textbf{92.23}\\ 
    \bottomrule
    \end{tabu}
    \caption{F1 on English NER tasks. Both sides of the entanglement model outperform the corresponding backbone models, and the subword side outperforms $\text{RoBERTa}_{\text{large}}$ (which has more parameters) and CharBERT. 
    \#C means the number of co-attention modules. 
    }
    \label{tab:noisy_text}
\end{table}



Table \ref{tab:noisy_text} shows the results of our model on two English NER datasets: WNUT-17 (noisy text) and CONLL-03 (formal text). Across all experiments, our model consistently outperforms the backbone models on both the subword and character sides. Interestingly, the improvement is more pronounced for WNUT-17 compared to CONLL-03, indicating that our model excels at handling noisy text. Additionally, we observe that the character side exhibits a more significant improvement than the subword side, suggesting that the character model benefits greatly from co-attending with the subword model. Although our models do not surpass the state-of-the-art (SOTA) performance, it is important to note that the SOTA models either rely on external context (CL-KL), employ neural architecture search across a broader range of models (ACE), or a linear chain CRF layer (ACE), making them less directly comparable to our model. 
Table \ref{tab:TweeBank} showcases the results of our model on TweeBank. Overall, we observe minimal differences between the entanglement model and the RoBERTa baseline. 

\begin{table}[t]
    \centering
    \begin{tabu}{l c c | c}
    \toprule
        \multicolumn{2}{l}{\textbf{Model}}   &  \textbf{TweeBank} & \textbf{WNUT-20} \\\midrule
        \rowfont{\color{gray}}
        \multicolumn{2}{l}{$\text{BERTweet}$}  & 95.20 & - \\
        \rowfont{\color{gray}}
        \multicolumn{2}{l}{NutCracker} & -  & 90.96$^\dagger$ \\
        \multicolumn{2}{l}{CharBERT} & 93.59 & 88.08 \\
        \multicolumn{2}{l}{$\text{RoBERTa}_{\text{base}}$} & 95.41 & 88.93 \\
        \multicolumn{2}{l}{$\text{RoBERTa}_{\text{large}}$} & 94.50 & 89.21 \\
        
        \midrule \midrule
        \textbf{Side} & \textbf{\#C} \\ \midrule
        
        \wordside  & 1 & 95.39 & 89.14 \\
         & 2  & \textbf{95.52}$^\dagger$ & \textbf{89.98} \\
         & 3 & 95.42 & 88.86 \\
         \bottomrule
    \end{tabu}
    \caption{Accuracy on TweeBank and F1 on WNUT-20. The entanglement model outperforms $\text{RoBERTa}_{\text{base}}$, $\text{RoBERTa}_{\text{large}}$, and CharBERT on these tasks. \#C refers to the number of co-attention modules.
    }
    \label{tab:TweeBank}
\end{table}







\subsection{Multilingual NER}

\begin{table*}[t]
    \centering
    \resizebox{\textwidth}{!}{
    \begin{tabu}{l c c c c c c c c c  c c |c  }
    
        \toprule

        \multicolumn{2}{l}{\textbf{Model}} & \textbf{AMH} & \textbf{HAU} & \textbf{IBO} & \textbf{KIN} & \textbf{LUG} & \textbf{LUO} & \textbf{PCM} & \textbf{SWA} & \textbf{WOL} & \textbf{YOR} & \textbf{Avg} \\
        
        \midrule
        \rowfont{\color{gray}}
         \multicolumn{2}{l}{PIXEL} & 47.7  & 82.4  & 79.9  & 64.2  & 76.5 & 66.6 & 78.7 & 79.8  & 59.7  & 70.7 & 70.62 \\
         \rowfont{\color{gray}}
          \multicolumn{2}{l}{ $\text{CANINE-c}_{\text{+n-grams}}$} & 50.0  & 88.0  & 85.0 & 72.8 & 79.6   & 74.2  & 88.7 & 83.7  & 66.5 $^\dagger$ & 79.1   & 76.76 \\
          
         \multicolumn{2}{l}{CANINE-s} & 32.70 & 74.38 & 71.79 & 55.92 & 69.98 & 53.75 & 66.17 & 73.37 & 57.82 & 61.00 & 61.69          \\
         \multicolumn{2}{l}{$\text{XLM-R}_{\text{base}}$} & 71.69 & \text{90.05} & 84.79 & 73.35 & 78.33 & 73.98 & 87.96 & 86.46 & 63.43 &\text{ 77.56} & 78.76 \\

         \multicolumn{2}{l}{$\text{XLM-R}_{\text{large}}$} & {75.51} & \text{91.06} & 83.85 & \textbf{76.61}$^\dagger$ & 78.09 & \textbf{77.08}$^\dagger$ & 	\text{90.08} & \text{88.87} & 65.58 &\text{ 79.50} & \text{80.62} \\

        \midrule \midrule
        \textbf{Side}& \textbf{\#C} \\ \midrule
       
        \charside  &1  & 41.99 & 79.12 & 74.00 & 59.23 & 70.48 & 61.17 & 75.06 & 78.25 & 59.19 & 63.45 & 66.20 \\
        &2  & 39.17 & 78.33 & 74.48 & 58.50 & 69.02 & 56.20 & 74.50 & 77.24 & 53.11 & 61.78 & 64.24          \\
        &3  & 41.14 & 79.00 & 73.81 & 58.89 & 70.53 & 55.56 & 73.67 & 77.58 & 57.71 & 59.67 & 64.76  \\
        
        \midrule

        \wordside&1  & 70.44 & 89.66 & {85.17} & 73.65 & 77.76 & 75.88 & 87.74 & 87.35 & 64.73 & 76.35 & 78.87 \\
        
        & 2  & \text{72.83}& 89.89 & 84.71 & 72.53 & \text{78.44} & \text{75.94} & \text{88.01} & 86.54 & \text{65.66} & 77.25 & \text{79.18} \\
        
        & 3  & 71.79 & 89.45 & 84.38 & \text{73.86} & 77.03 & 74.60 & 87.61 & \text{87.39} & 64.77 & 76.76 & 78.76   \\

        \midrule 

        \wordside &1  & 74.01 & 91.35 & 84.33 & 74.83 & 79.08 & 75.89 & \textbf{90.60}$^\dagger$ & \textbf{89.58}$^\dagger$ &65.40  & 77.81 & 80.29 \\
        
        ($\text{XLM-R}_{\text{large}}$)& 2  & \text{74.14}& 90.67 & 85.03 &  72.52& 79.93 & 75.40  & 90.10  &  89.62 & \textbf{66.13} & 78.29 & 80.18 \\
        
        & 3  & \textbf{76.67}$^\dagger$ & \textbf{91.90}$^\dagger$ & \textbf{85.83}$^\dagger$ & 73.42 & \textbf{80.16}$^\dagger$ & 75.24 &88.98  & 88.60 & 65.82 & \textbf{80.49}$^\dagger$  &  \textbf{80.71}$^\dagger$\\
        \bottomrule
    \end{tabu}
    }
    \caption{MasakhaNER F1 score for Multilingual NER results. The first 2 panels (\charside, \wordside) refers to the two sides of EM trained with XLM-R as the backbone. The bottom panel utilizes EM with $\text{XLM-R}_{\text{large}}$ as the backbone. Both sides of the best entanglement model consistently outperform the corresponding backbone models ($\text{XLM-R}_{\text{base}}$ and CANINE-S). EM with $\text{XLM-R}_{\text{large}}$ as the subword backbone archives SOTA performance on 6 out of 10 languages. 
    \#C means the number of co-attention modules.
    The last column Avg indicates the macro average F1 score of all the 10 African languages. 
    }
    \label{tab:masak}
\end{table*}

\begin{table}[t]
    \centering
    \begin{tabu}{l c c  c c c} 
    \toprule
        \textbf{Evaluation} & \textbf{All} & \textbf{All} & \textbf{MIX} & \textbf{MIX} \\ \midrule
        \textbf{Data}  &  \textbf{S-W} & \textbf{G-T}  &  \textbf{S-W} & \textbf{G-T} \\\midrule
        \rowfont{\color{gray}}
        SegRNN  & $92.40^\dagger$ & 93.60 & 84.6& 72.9 \\
        CANINE-s& 90.84 & 94.12 &82.97 & 72.44 \\
        
        \midrule \midrule
        \textbf{Side-\#C}  \\ \midrule
        
        \charside-1 & \textbf{91.24} &94.60& $\textbf{86.23}^\dagger$ & $\textbf{74.21}^\dagger$ \\
         \charside-2  & 91.17 &94.74 & 84.42 & $73.82$ \\
         \charside-3 & 91.00 & 94.39 & 84.05 & 71.26\\
         \charside-4 & 91.00 &$\textbf{94.86}^\dagger$ &84.42 & 72.63

\\
         \bottomrule
    \end{tabu}
    \caption{Character accuracy on code-switching tasks. The entanglement model outperforms CANINE-s and previous studies across all sub-tasks, and it outperforms SegRNN \citep{mager2019subwordlevel} except (All, S-W).  ``All" means the accuracy of all data, and ``MIX" means the accuracy of words with intraword switching. S-W refers to Spanish-Wixarica, G-T refers to German-Turkish. 
    }
    \label{tab:codeswitch}
    \vspace{-2ex}
\end{table}

%
The results of our model on MasakhaNER are presented in Table \ref{tab:masak}. Again, we observe that our model outperforms the baseline models on both the subword and character side, with a more substantial improvement on the character side. The performance boost for certain languages, such as Luo (LUO) and Wolof (WOL), appears more substantial. Luo consists of additional consonants and nine vowels \citep{adelani-etal-2021-masakhaner}, which might be better processed by the character model. Wolof's morphology is derivationally rich \cite{ka_wolof_1987}, which may suggest that our model performs better on morphologically rich languages because it effectively leverages the character model.  

Motivated by the performance gap between XLM-R and its larger variant, $\text{XLM-R}_{\text{large}}$, we experimented with the entanglement model using $\text{XLM-R}_{\text{large}}$ as the foundational subword backbone. To reconcile the embedding dimension mismatch between the two backbones (768 for CANINE-S and 1024 for $\text{XLM-R}_{\text{large}}$), we employed a fully-connected linear layer to upscale CANINE's character embeddings before passing them to the co-attention layers. As illustrated in the bottom panel of Table \ref{tab:masak}, when the entanglement model utilize $\text{XLM-R}_{\text{large}}$ as the backbone, its performance surpasses the standalone $\text{XLM-R}_{\text{large}}$ model, and it archives SOTA performance across most languages.

\subsection{Character-level Sequence Labeling}
Table \ref{tab:codeswitch} shows the results of our model on intraword code-switching tasks. We see that the entanglement model outperforms CANINE-s across all tasks and specifications, and it outperforms the previous SOTA SegRNN \citep{mager2019subwordlevel} in most tasks. The performance gain is more substantial for ``Mixed" words, which contain intraword code-witching.

\subsection{Classification}

Table \ref{tab:TweetEval} presents the results of our model on the WNUT-2020 shared task \#2, demonstrating its superiority over the baseline RoBERTa model and achieving performance close to state-of-the-art (the NutCracker model \cite{kumar-singh-2020-nutcracker}).

Furthermore, Table \ref{tab:TweetEval} showcases the results of the TweetEval benchmark, where our model outperforms the backbone models that have not been pretrained on this type of noisy text. For some subtasks, our performance is competitive with BERTweet \cite{nguyen-etal-2020-bertweet}, which is pretrained on Twitter text. We also observe that the improvement on the character side is more substantial than the subword side.


\begin{table*}[t]
\centering 
\begin{tabu}{l c c c c c c c | c}
\toprule
\multicolumn{2}{l}{\textbf{Model}} &  \textbf{Emoji}          & \textbf{Emotion}        & \textbf{Hate}          & \textbf{Irony}          & \textbf{Offsensive}     & \textbf{Sentiment}    & \textbf{Avg}  \\  
\midrule

\rowfont{\color{gray}}
\multicolumn{2}{l}{RoB-RT} & 31.4 & 79.5$^\dagger$   & 52.3  & 61.7   & 80.5  & 72.6   & 63.00 \\

\multicolumn{2}{l}{BERTweet} & 33.58& 78.88&\textbf{53.87}&\textbf{80.53}&80.17&68.53&\textbf{65.93}
 \\

\multicolumn{2}{l}{CANINE-s}& 26.27          & 61.72          & 43.51         & 61.96          & 73.63          & 61.72        & 54.80  \\
\multicolumn{2}{l}{$\text{RoBERTa}_{\text{base}}$} & 33.36          & 78.55          & 50.49         & 73.14          & 78.05          & 68.28     &   63.65  \\
\multicolumn{2}{l}{$\text{RoBERTa}_{\text{large}}$} & 34.25          & \textbf{81.87}        & 51.08        & 70.75          & 80.29          & 71.40    &   64.94  \\

\multicolumn{2}{l}{CharBERT} & 30.68          & 75.56          & 48.11         & 68.72          & 70.95         & \textbf{71.62}   &   60.94  \\

\midrule\midrule

 \textbf{Side} & \textbf{\#C} \\\midrule

\charside 
 & 1 & 31.47          & 66.43          & 46.25         & 69.38          & \textbf{81.24} $^\dagger$ & 70.39     &   60.86   \\
 & 2 & 31.00             & 77.46          & 50.05         & 67.46          & 80.41          & 70.86        & 62.87  \\ 
 \midrule
\wordside   & 1 & 33.56           & \text{79.31} & 50.19         & {73.95 }         & 80.57          & 70.62        &  64.70 \\
  & 2 & \textbf{34.38} $^\dagger$         & 78.65          & {52.60} & 73.69          & 80.02          & \text{71.41 }    &  {65.13} \\ 
(Pretrain)& 1 & 30.33 & 	74.02 & 	44.81 &	59.87	& 78.27	& 66.42	& 58.95 \\
  \bottomrule
\end{tabu}
\caption{F1 on TweetEval. Both sides of the best entanglement model (EM) outperform the corresponding backbone models ($\text{XLM-R}_{\text{base}}$ and CANINE-S) and CharBERT across all tasks except Sentiment.  For all models we have evaluated (not including BERTweet, which is pre-trained on Twitter text), EM performs the best for 4 out of 6 subtasks. 
The last column Avg indicates the macro average F1 across 6 tasks. }
\label{tab:TweetEval}    
\end{table*}

\subsection{Discussion}

\paragraph{Character Models}

In most tasks, we see that the performance of CANINE-s is not comparable with RoBERTa. This perhaps explains the observation that the improvement of our model on the character is usually much more substantial than the subword side. Thus, our model might benefit from a different (potentially stronger) character model, such as Charformer \cite{tay2022charformer} and ByT5 \cite{xue-etal-2022-byt5}, and we leave it for future research. 

\paragraph{Number of Co-attention Modules} 





Generally, we observe that using two co-attention modules appears to be the optimal choice for the subword side, while one co-attention module appears to suffice for the character side. Although in certain tasks using 4 co-attention modules yields the highest performance, these additional benefits of more co-attention modules appear minimal.

\paragraph{Efficiency}
\label{sec:parameter-size}

Our entanglement model requires 2-3 times the memory of a single backbone model. The 2-COTRM entanglement model contains around 290M parameters, whereas its subword backbone ($\text{RoBERTa}_{\text{base}}$) contains 125M parameters. Yet, our model contains fewer parameters than larger pre-trained models like $\text{RoBERTa}_{\text{large}}$ (354M parameters). Our model has higher parallelizability than $\text{RoBERTa}_{\text{large}}$, as the computation of the character and subword model is independent before the co-attention module. Empirically, the runtime of the entanglement model is roughly 1.72 times of $\text{RoBERTa}_{\text{base}}$ and 0.54 times of $\text{RoBERTa}_{\text{large}}$. 

\paragraph{Baseline}

Table \ref{tab:noisy_text}, \ref{tab:TweeBank}, \ref{tab:TweetEval} shows that the entanglement model outperforms $\text{RoBERTa}_{\text{large}}$, which is pretrained and has more parameters, across all English classification and sequence labeling tasks. Table \ref{tab:masak} shows that the entanglement model with XLM-R as the backbone failed to outperform $\text{XLM-R}_{\text{large}}$, so maybe more pretraining is required for lower-resourced languages. We see that EM with $\text{XLM-R}_{\text{large}}$ as the backbone outperforms $\text{XLM-R}_{\text{large}}$. Also, the entanglement model outperforms CharBERT (an English-only model) across all English classification and sequence labeling tasks, suggesting that our model more effectively leverages the ability of both character and subword models. 





\section{Model Extensions}

In this section, we explore two natural extensions that demonstrate how the simplicity of our model eliminates the need for additional complexity.




\paragraph{Positional Embeddings}

We experimented with several ways to add positional embeddings (PE) in the co-attention module. Details on PE training are in appendix \ref{app:positional-embeddings}. 
From table \ref{tab:pos_embed_result}, we see that 
for WNUT-17, adding PEs hurts the model's performance. In CONLL-03, strategy C has a slight improvement in the model's performance, though it appears very marginal. This suggests that the entanglement model autonomously learns the translation between subword PEs and character PEs.

\paragraph{MLM pretraining}

We pretrain a 1-layer entanglement model on 8\% of WikiText-103 \cite{merity2016pointer} and Bookcorpus \cite{zhu2015aligning}. Details on pretraining are in appendix \ref{app:pretraining}. 
From table \ref{tab:comparison} \& \ref{tab:TweetEval}, we see that the pre-trained model fails to outperform the standard, un-pretrained entanglement model. This suggests that pretraining does not appear to help the model generalize. Nevertheless, it is also possible that the scale of pretraining is not large enough for it to exhibit a positive influence, and we leave it to future work.

\begin{table}[t]
\centering
    \begin{tabular}{c  c  c}
    \toprule
        \textbf{Strategy} & \textbf{WNUT-17} & \textbf{CONLL-03} \\ 
        \midrule
        No PEs & \textbf{57.80} & 91.81\\
        A & 56.23 & 91.88 \\
        B & 57.06 & 91.79 \\
        C & 56.94 & \textbf{92.15} \\
        \bottomrule
    \end{tabular}
    \caption{NER F1 results for different kinds of positional embeddings (PEs). For WNUT-17, adding PEs decreases model performance. For CONLL-03, adding PE A and C leads to a very marginal performance boost. }
    \label{tab:pos_embed_result}
\end{table}





\begin{table}[t]
    \centering
    \begin{tabular}{l l l l ||l   }
    \toprule
    \textbf{Dataset}& RL & CB & EM-P & EM \\ \midrule
    TweeBank &94.50 & 93.59 &	93.97	& \textbf{95.52} \\
    WNUT-20 & 89.21 &	88.08	&88.08	&\textbf{89.98} \\
    WNUT-17 & 	56.38	&53.63&	51.71&	\textbf{57.97} \\
    CONLL-03 & 	92.31 &92.07 &91.21 &	\textbf{92.23} \\
    \bottomrule 
    \end{tabular}
    \caption{A more direct comparison between EM and larger pre-trained models (RoBERTa$_\text{large}$, RL), another character-aware subword model (CharBERT, CB), and pre-trained EM. The standard EM outperforms these three models across all these four tasks.}
    \label{tab:comparison}
\end{table}

\section{Related Work}

Many existing studies have investigated learning subward representations from multiple granularities of input (\S\ref{sec:input-granularities}), and many studies has explored learning character representations (\S\ref{sec:learning_char_rep}).
Comparatively few works have explored outputting representations at multiple granularities (\S\ref{sec:output-granularities}). 
Our model draws inspiration from studies in multimodal machine learning (\S\ref{sec:multimodal-nlp}) and facilitates information exchange between subword and character representations through a co-attention module.

\subsection{Multiple Granularities of Input}
\label{sec:input-granularities}
Several previous studies have explored the use of multiple granularities in input representation. Charformer \citep{tay2022charformer} uses a data-driven method to learn subword representation from characters. CharBERT \citet{ma-etal-2020-charbert} learns two \emph{subword-level} representations, respectively containing subword and character-level information. The ACE model \citep{wang-etal-2021-automated} employs neural architecture search to determine the optimal combination of embeddings. \citet{sanh_hierarchical_2019} merge embeddings from various text granularities before inputting them into the encoder \citep{sanh_hierarchical_2019}.  \citet{shahzad-etal-2021-inferner} and \citet{aguilar-etal-2018-modeling} employing separate encoders to extract contextualized representations for different granularities of text, which are later combined during inference. All these studies produce subword-level representations, but they produce no useful representations for other text granularities.

\subsection{Character Representation Learning}
\label{sec:learning_char_rep}
Models like CANINE \citep{clark-etal-2022-canine} and ByT5 \citep{xue-etal-2022-byt5} directly pre-train a character-level transformer to obtain character representations. However, character models could be hard to train as they assume less structure about the text. To mitigate this issue, \citet{sun-etal-2023-characters} uses a hierarchical structure to integrate word boundary information in the character model. \citet{huang-etal-2023-inducing} learns character representation inside a subword model by treating characters as type variables in a causal model. Studies found that incorporating linguistic features of the characters, such as phonetic information \citep{matsuhira2023ipaclip}, Chinese character shape and Pinyin \citep{sun-etal-2021-chinesebert, wei-etal-2023-ptcspell}, can yield performance gains. 

\subsection{Multiple Granularities of Output}
\label{sec:output-granularities}

In contrast to the extensive research on processing multiple granularities as input, there have been limited studies proposing models that generate multiple granularities of output. In speech recognition, \citet{sanabria_hierarhical_2018} train a single model to simultaneously produce text transcripts at different granularities, specifically characters, and subwords with varying vocabulary sizes. \citet{https://doi.org/10.48550/arxiv.1910.12368} employs a shared encoder but separate decoders for different output granularities, allowing decoders to generate outputs concurrently. \citet{https://doi.org/10.48550/arxiv.1812.02308}optimize different models for distinct granularities of text jointly, using a combined loss.

\subsection{Multimodal NLP}
\label{sec:multimodal-nlp}
Prior research has demonstrated the potential benefits of incorporating non-linguistic modalities in various NLP tasks. For instance, ChineseBERT \citep{sun-etal-2021-chinesebert} incorporates Pinyin and glyph information of Chinese characters during pretraining, leading to performance boosts in Chinese NLP tasks. 
Our work draws inspiration from vision-language models. 
Models like VisualBERT \citep{li2019visualbert} and VL-BERT \citep{su2020vlbert} learn a shared representation space for both images and language, utilizing a single transformer as the encoder for both modalities. In contrast, our model utilizes pretrained subword and character models and employs the co-attention module, as adopted by ViLBERT \citep{NEURIPS2019_c74d97b0}, to facilitate information exchange between the two granularities.

\section{Conclusion}
In this paper, we introduce a novel entanglement model to effectively combine character and subword language models using co-attention modules. Unlike many prior works, our model produces mutually informed representations of subwords and characters, which could be used to process both subword and character-level tasks. Its architecture is model-agnostic, and it opens new directions for pretraining and scaling up. Empirically, our model has demonstrated improvements over the baseline models on various sequence labeling and text classification tasks. 
Our entanglement model achieves state-of-the-art results on various tasks/settings: POS tagging on TweeBank, NER on Ibo and Wolof from MasakhaNER, and intraword code-switching on German-Turkish. 
Notably, the improvement of our model is most significant for noisy texts and low-resourced, morphologically rich languages. Furthermore, the entanglement model outperforms larger pretrained subword models, which have higher parameter counts, on most tasks.  
%
While our model features a simple architecture, incorporating extensions like positional embeddings or additional pretraining do not improve its performance, which implies that the model's structure facilitates the learning of relevant information during fine-tuning, rendering additional complexities unnecessary.




\section{Limitations}
\paragraph{Computational Efficiency} Although our model demands greater computational resources and may have a slower optimization process compared to the backbone model (e.g., RoBERTa), it still faster than previous models like ACE, which utilize multiple embeddings from different models through neural architecture search. Moreover, in \S \ref{sec:parameter-size}, we demonstrate that the performance of our model is comparable, if not superior, to the $\text{RoBERTa}_{\text{large}}$ model, which has a higher parameter count.

\paragraph{Model Extension} 
Our model is designed to accommodate a maximum of two backbone models, and there is no straightforward way to extend it for the utilization of three or more backbone models. While exploring the entanglement model with three backbone models could be an intriguing avenue for researchers interested in word-level modeling, it's worth noting that the majority of current language modeling primarily focuses on the two levels (character or subword) employed in our entanglement models.

\paragraph{Pretraining} 
During the pretraining phase, we observed a rapid decrease in the character MLM loss compared to the subword MLM loss. To introduce more challenging training objectives, one option is to mask out an entire word's worth of characters instead of just a single character at a time. This strategy could potentially encourage the model to capture more nuanced details in the text, leading to potential performance improvements. Additionally, due to limitations in computational resources, we performed pretraining on a subset (approximately 8\%) of the corpus instead of conducting a full-scale pretraining. As such, further investigation into this approach is left to future research.





\begin{thebibliography}{41}
\expandafter\ifx\csname natexlab\endcsname\relax\def\natexlab#1{#1}\fi

\bibitem[{Adelani et~al.(2021)Adelani, Abbott, Neubig, D{'}souza, Kreutzer, Lignos, Palen-Michel, Buzaaba, Rijhwani, Ruder, Mayhew, Azime, Muhammad, Emezue, Nakatumba-Nabende, Ogayo, Anuoluwapo, Gitau, Mbaye, Alabi, Yimam, Gwadabe, Ezeani, Niyongabo, Mukiibi, Otiende, Orife, David, Ngom, Adewumi, Rayson, Adeyemi, Muriuki, Anebi, Chukwuneke, Odu, Wairagala, Oyerinde, Siro, Bateesa, Oloyede, Wambui, Akinode, Nabagereka, Katusiime, Awokoya, MBOUP, Gebreyohannes, Tilaye, Nwaike, Wolde, Faye, Sibanda, Ahia, Dossou, Ogueji, DIOP, Diallo, Akinfaderin, Marengereke, and Osei}]{adelani-etal-2021-masakhaner}
David~Ifeoluwa Adelani, Jade Abbott, Graham Neubig, Daniel D{'}souza, Julia Kreutzer, Constantine Lignos, Chester Palen-Michel, Happy Buzaaba, Shruti Rijhwani, Sebastian Ruder, Stephen Mayhew, Israel~Abebe Azime, Shamsuddeen~H. Muhammad, Chris~Chinenye Emezue, Joyce Nakatumba-Nabende, Perez Ogayo, Aremu Anuoluwapo, Catherine Gitau, Derguene Mbaye, Jesujoba Alabi, Seid~Muhie Yimam, Tajuddeen~Rabiu Gwadabe, Ignatius Ezeani, Rubungo~Andre Niyongabo, Jonathan Mukiibi, Verrah Otiende, Iroro Orife, Davis David, Samba Ngom, Tosin Adewumi, Paul Rayson, Mofetoluwa Adeyemi, Gerald Muriuki, Emmanuel Anebi, Chiamaka Chukwuneke, Nkiruka Odu, Eric~Peter Wairagala, Samuel Oyerinde, Clemencia Siro, Tobius~Saul Bateesa, Temilola Oloyede, Yvonne Wambui, Victor Akinode, Deborah Nabagereka, Maurice Katusiime, Ayodele Awokoya, Mouhamadane MBOUP, Dibora Gebreyohannes, Henok Tilaye, Kelechi Nwaike, Degaga Wolde, Abdoulaye Faye, Blessing Sibanda, Orevaoghene Ahia, Bonaventure F.~P. Dossou, Kelechi Ogueji, Thierno~Ibrahima DIOP,
  Abdoulaye Diallo, Adewale Akinfaderin, Tendai Marengereke, and Salomey Osei. 2021.
\newblock \href {https://doi.org/10.1162/tacl_a_00416} {{M}asakha{NER}: Named entity recognition for {A}frican languages}.
\newblock \emph{Transactions of the Association for Computational Linguistics}, 9:1116--1131.

\bibitem[{Aguilar et~al.(2018)Aguilar, L{\'o}pez-Monroy, Gonz{\'a}lez, and Solorio}]{aguilar-etal-2018-modeling}
Gustavo Aguilar, Adrian~Pastor L{\'o}pez-Monroy, Fabio Gonz{\'a}lez, and Thamar Solorio. 2018.
\newblock \href {https://doi.org/10.18653/v1/N18-1127} {Modeling noisiness to recognize named entities using multitask neural networks on social media}.
\newblock In \emph{Proceedings of the 2018 Conference of the North {A}merican Chapter of the Association for Computational Linguistics: Human Language Technologies, Volume 1 (Long Papers)}, pages 1401--1412, New Orleans, Louisiana. Association for Computational Linguistics.

\bibitem[{Barbieri et~al.(2020)Barbieri, Camacho-Collados, Espinosa~Anke, and Neves}]{barbieri-etal-2020-tweeteval}
Francesco Barbieri, Jose Camacho-Collados, Luis Espinosa~Anke, and Leonardo Neves. 2020.
\newblock \href {https://doi.org/10.18653/v1/2020.findings-emnlp.148} {{T}weet{E}val: Unified benchmark and comparative evaluation for tweet classification}.
\newblock In \emph{Findings of the Association for Computational Linguistics: EMNLP 2020}, pages 1644--1650, Online. Association for Computational Linguistics.

\bibitem[{Clark et~al.(2022)Clark, Garrette, Turc, and Wieting}]{clark-etal-2022-canine}
Jonathan~H. Clark, Dan Garrette, Iulia Turc, and John Wieting. 2022.
\newblock \href {https://doi.org/10.1162/tacl_a_00448} {Canine: Pre-training an efficient tokenization-free encoder for language representation}.
\newblock \emph{Transactions of the Association for Computational Linguistics}, 10:73--91.

\bibitem[{Conneau et~al.(2020)Conneau, Khandelwal, Goyal, Chaudhary, Wenzek, Guzm{\'a}n, Grave, Ott, Zettlemoyer, and Stoyanov}]{conneau-etal-2020-unsupervised}
Alexis Conneau, Kartikay Khandelwal, Naman Goyal, Vishrav Chaudhary, Guillaume Wenzek, Francisco Guzm{\'a}n, Edouard Grave, Myle Ott, Luke Zettlemoyer, and Veselin Stoyanov. 2020.
\newblock \href {https://doi.org/10.18653/v1/2020.acl-main.747} {Unsupervised cross-lingual representation learning at scale}.
\newblock In \emph{Proceedings of the 58th Annual Meeting of the Association for Computational Linguistics}, pages 8440--8451, Online. Association for Computational Linguistics.

\bibitem[{Derczynski et~al.(2017)Derczynski, Nichols, van Erp, and Limsopatham}]{derczynski-etal-2017-results}
Leon Derczynski, Eric Nichols, Marieke van Erp, and Nut Limsopatham. 2017.
\newblock \href {https://doi.org/10.18653/v1/W17-4418} {Results of the {WNUT}2017 shared task on novel and emerging entity recognition}.
\newblock In \emph{Proceedings of the 3rd Workshop on Noisy User-generated Text}, pages 140--147, Copenhagen, Denmark. Association for Computational Linguistics.

\bibitem[{Devlin et~al.(2019)Devlin, Chang, Lee, and Toutanova}]{devlin-etal-2019-bert}
Jacob Devlin, Ming-Wei Chang, Kenton Lee, and Kristina Toutanova. 2019.
\newblock \href {https://doi.org/10.18653/v1/N19-1423} {{BERT}: Pre-training of deep bidirectional transformers for language understanding}.
\newblock In \emph{Proceedings of the 2019 Conference of the North {A}merican Chapter of the Association for Computational Linguistics: Human Language Technologies, Volume 1 (Long and Short Papers)}, pages 4171--4186, Minneapolis, Minnesota. Association for Computational Linguistics.

\bibitem[{Huang et~al.(2023)Huang, Wu, Mahowald, and Potts}]{huang-etal-2023-inducing}
Jing Huang, Zhengxuan Wu, Kyle Mahowald, and Christopher Potts. 2023.
\newblock \href {https://doi.org/10.18653/v1/2023.findings-acl.770} {Inducing character-level structure in subword-based language models with type-level interchange intervention training}.
\newblock In \emph{Findings of the Association for Computational Linguistics: ACL 2023}, pages 12163--12180, Toronto, Canada. Association for Computational Linguistics.

\bibitem[{Jiang et~al.(2022)Jiang, Hua, Beeferman, and Roy}]{jiang-etal-2022-annotating}
Hang Jiang, Yining Hua, Doug Beeferman, and Deb Roy. 2022.
\newblock \href {https://aclanthology.org/2022.lrec-1.780} {Annotating the {T}weebank corpus on named entity recognition and building {NLP} models for social media analysis}.
\newblock In \emph{Proceedings of the Thirteenth Language Resources and Evaluation Conference}, pages 7199--7208, Marseille, France. European Language Resources Association.

\bibitem[{Ka(1987)}]{ka_wolof_1987}
Omar Ka. 1987.
\newblock \href {https://www.proquest.com/docview/303589214/abstract/D8F7FE3811F44744PQ/1} {\emph{Wolof phonology and morphology: {A} non-linear approach}}.
\newblock Ph.{D}., University of Illinois at Urbana-Champaign, United States -- Illinois.
\newblock ISBN: 9798206799132.

\bibitem[{Kremer et~al.(2018)Kremer, Borgholt, and Maaløe}]{https://doi.org/10.48550/arxiv.1812.02308}
Jan Kremer, Lasse Borgholt, and Lars Maaløe. 2018.
\newblock \href {https://doi.org/10.48550/ARXIV.1812.02308} {On the inductive bias of word-character-level multi-task learning for speech recognition}.

\bibitem[{Kudo and Richardson(2018)}]{kudo-richardson-2018-sentencepiece}
Taku Kudo and John Richardson. 2018.
\newblock \href {https://doi.org/10.18653/v1/D18-2012} {{S}entence{P}iece: A simple and language independent subword tokenizer and detokenizer for neural text processing}.
\newblock In \emph{Proceedings of the 2018 Conference on Empirical Methods in Natural Language Processing: System Demonstrations}, pages 66--71, Brussels, Belgium. Association for Computational Linguistics.

\bibitem[{Kumar and Singh(2020)}]{kumar-singh-2020-nutcracker}
Priyanshu Kumar and Aadarsh Singh. 2020.
\newblock \href {https://doi.org/10.18653/v1/2020.wnut-1.57} {{N}ut{C}racker at {WNUT}-2020 task 2: Robustly identifying informative {COVID}-19 tweets using ensembling and adversarial training}.
\newblock In \emph{Proceedings of the Sixth Workshop on Noisy User-generated Text (W-NUT 2020)}, pages 404--408, Online. Association for Computational Linguistics.

\bibitem[{Li et~al.(2019)Li, Yatskar, Yin, Hsieh, and Chang}]{li2019visualbert}
Liunian~Harold Li, Mark Yatskar, Da~Yin, Cho-Jui Hsieh, and Kai-Wei Chang. 2019.
\newblock \href {http://arxiv.org/abs/1908.03557} {Visualbert: A simple and performant baseline for vision and language}.

\bibitem[{Lu et~al.(2019{\natexlab{a}})Lu, Batra, Parikh, and Lee}]{lu2019vilbert}
Jiasen Lu, Dhruv Batra, Devi Parikh, and Stefan Lee. 2019{\natexlab{a}}.
\newblock \href {http://arxiv.org/abs/1908.02265} {Vilbert: Pretraining task-agnostic visiolinguistic representations for vision-and-language tasks}.

\bibitem[{Lu et~al.(2019{\natexlab{b}})Lu, Batra, Parikh, and Lee}]{NEURIPS2019_c74d97b0}
Jiasen Lu, Dhruv Batra, Devi Parikh, and Stefan Lee. 2019{\natexlab{b}}.
\newblock \href {https://proceedings.neurips.cc/paper_files/paper/2019/file/c74d97b01eae257e44aa9d5bade97baf-Paper.pdf} {Vilbert: Pretraining task-agnostic visiolinguistic representations for vision-and-language tasks}.
\newblock In \emph{Advances in Neural Information Processing Systems}, volume~32. Curran Associates, Inc.

\bibitem[{Ma et~al.(2020)Ma, Cui, Si, Liu, Wang, and Hu}]{ma-etal-2020-charbert}
Wentao Ma, Yiming Cui, Chenglei Si, Ting Liu, Shijin Wang, and Guoping Hu. 2020.
\newblock \href {https://doi.org/10.18653/v1/2020.coling-main.4} {{C}har{BERT}: Character-aware pre-trained language model}.
\newblock In \emph{Proceedings of the 28th International Conference on Computational Linguistics}, pages 39--50, Barcelona, Spain (Online). International Committee on Computational Linguistics.

\bibitem[{Mager et~al.(2019)Mager, Özlem Çetinoğlu, and Kann}]{mager2019subwordlevel}
Manuel Mager, Özlem Çetinoğlu, and Katharina Kann. 2019.
\newblock \href {http://arxiv.org/abs/1904.01989} {Subword-level language identification for intra-word code-switching}.

\bibitem[{Matsuhira et~al.(2023)Matsuhira, Kastner, Komamizu, Hirayama, Doman, Kawanishi, and Ide}]{matsuhira2023ipaclip}
Chihaya Matsuhira, Marc~A. Kastner, Takahiro Komamizu, Takatsugu Hirayama, Keisuke Doman, Yasutomo Kawanishi, and Ichiro Ide. 2023.
\newblock \href {http://arxiv.org/abs/2303.03144} {Ipa-clip: Integrating phonetic priors into vision and language pretraining}.

\bibitem[{Merity et~al.(2016)Merity, Xiong, Bradbury, and Socher}]{merity2016pointer}
Stephen Merity, Caiming Xiong, James Bradbury, and Richard Socher. 2016.
\newblock \href {http://arxiv.org/abs/1609.07843} {Pointer sentinel mixture models}.

\bibitem[{Nguyen et~al.(2020{\natexlab{a}})Nguyen, Vu, Rahimi, Dao, Nguyen, and Doan}]{nguyen2020wnut2020}
Dat~Quoc Nguyen, Thanh Vu, Afshin Rahimi, Mai~Hoang Dao, Linh~The Nguyen, and Long Doan. 2020{\natexlab{a}}.
\newblock \href {http://arxiv.org/abs/2010.08232} {Wnut-2020 task 2: Identification of informative covid-19 english tweets}.

\bibitem[{Nguyen et~al.(2020{\natexlab{b}})Nguyen, Vu, and Tuan~Nguyen}]{nguyen-etal-2020-bertweet}
Dat~Quoc Nguyen, Thanh Vu, and Anh Tuan~Nguyen. 2020{\natexlab{b}}.
\newblock \href {https://doi.org/10.18653/v1/2020.emnlp-demos.2} {{BERT}weet: A pre-trained language model for {E}nglish tweets}.
\newblock In \emph{Proceedings of the 2020 Conference on Empirical Methods in Natural Language Processing: System Demonstrations}, pages 9--14, Online. Association for Computational Linguistics.

\bibitem[{Peters et~al.(2018)Peters, Neumann, Iyyer, Gardner, Clark, Lee, and Zettlemoyer}]{peters-etal-2018-deep}
Matthew~E. Peters, Mark Neumann, Mohit Iyyer, Matt Gardner, Christopher Clark, Kenton Lee, and Luke Zettlemoyer. 2018.
\newblock \href {https://doi.org/10.18653/v1/N18-1202} {Deep contextualized word representations}.
\newblock In \emph{Proceedings of the 2018 Conference of the North {A}merican Chapter of the Association for Computational Linguistics: Human Language Technologies, Volume 1 (Long Papers)}, pages 2227--2237, New Orleans, Louisiana. Association for Computational Linguistics.

\bibitem[{Sanabria and Metze(2018)}]{sanabria_hierarhical_2018}
Ramon Sanabria and Florian Metze. 2018.
\newblock \href {https://doi.org/10.48550/ARXIV.1807.07104} {Hierarchical multi task learning with ctc}.

\bibitem[{Sanh et~al.(2019)Sanh, Wolf, and Ruder}]{sanh_hierarchical_2019}
Victor Sanh, Thomas Wolf, and Sebastian Ruder. 2019.
\newblock A hierarchical multi-task approach for learning embeddings from semantic tasks.
\newblock In \emph{Proceedings of the {AAAI} {Conference} on {Artificial} {Intelligence}}, volume~33, pages 6949--6956.
\newblock Issue: 01.

\bibitem[{Sennrich et~al.(2016)Sennrich, Haddow, and Birch}]{sennrich-etal-2016-neural}
Rico Sennrich, Barry Haddow, and Alexandra Birch. 2016.
\newblock \href {https://doi.org/10.18653/v1/P16-1162} {Neural machine translation of rare words with subword units}.
\newblock In \emph{Proceedings of the 54th Annual Meeting of the Association for Computational Linguistics (Volume 1: Long Papers)}, pages 1715--1725, Berlin, Germany. Association for Computational Linguistics.

\bibitem[{Shahzad et~al.(2021)Shahzad, Amin, Esteves, and Ngonga~Ngomo}]{shahzad-etal-2021-inferner}
Moemmur Shahzad, Ayesha Amin, Diego Esteves, and Axel-Cyrille Ngonga~Ngomo. 2021.
\newblock \href {https://doi.org/10.32473/flairs.v34i1.128538} {Inferner: an attentive model leveraging the sentence-level information for named entity recognition in microblogs}.
\newblock In \emph{The international {FLAIRS} conference proceedings}, volume~34.

\bibitem[{Srinivasan et~al.(2019)Srinivasan, Sanabria, and Metze}]{https://doi.org/10.48550/arxiv.1910.12368}
Tejas Srinivasan, Ramon Sanabria, and Florian Metze. 2019.
\newblock \href {https://doi.org/10.48550/ARXIV.1910.12368} {Multitask learning for different subword segmentations in neural machine translation}.

\bibitem[{Su et~al.(2020)Su, Zhu, Cao, Li, Lu, Wei, and Dai}]{su2020vlbert}
Weijie Su, Xizhou Zhu, Yue Cao, Bin Li, Lewei Lu, Furu Wei, and Jifeng Dai. 2020.
\newblock \href {http://arxiv.org/abs/1908.08530} {Vl-bert: Pre-training of generic visual-linguistic representations}.

\bibitem[{Sun et~al.(2023)Sun, Luisier, Batmanghelich, Florencio, and Zhang}]{sun-etal-2023-characters}
Li~Sun, Florian Luisier, Kayhan Batmanghelich, Dinei Florencio, and Cha Zhang. 2023.
\newblock \href {https://doi.org/10.18653/v1/2023.acl-long.200} {From characters to words: Hierarchical pre-trained language model for open-vocabulary language understanding}.
\newblock In \emph{Proceedings of the 61st Annual Meeting of the Association for Computational Linguistics (Volume 1: Long Papers)}, pages 3605--3620, Toronto, Canada. Association for Computational Linguistics.

\bibitem[{Sun et~al.(2021)Sun, Li, Sun, Meng, Ao, He, Wu, and Li}]{sun-etal-2021-chinesebert}
Zijun Sun, Xiaoya Li, Xiaofei Sun, Yuxian Meng, Xiang Ao, Qing He, Fei Wu, and Jiwei Li. 2021.
\newblock \href {https://doi.org/10.18653/v1/2021.acl-long.161} {{C}hinese{BERT}: {C}hinese pretraining enhanced by glyph and {P}inyin information}.
\newblock In \emph{Proceedings of the 59th Annual Meeting of the Association for Computational Linguistics and the 11th International Joint Conference on Natural Language Processing (Volume 1: Long Papers)}, pages 2065--2075, Online. Association for Computational Linguistics.

\bibitem[{Tay et~al.(2022)Tay, Tran, Ruder, Gupta, Chung, Bahri, Qin, Baumgartner, Yu, and Metzler}]{tay2022charformer}
Yi~Tay, Vinh~Q. Tran, Sebastian Ruder, Jai Gupta, Hyung~Won Chung, Dara Bahri, Zhen Qin, Simon Baumgartner, Cong Yu, and Donald Metzler. 2022.
\newblock \href {http://arxiv.org/abs/2106.12672} {Charformer: Fast character transformers via gradient-based subword tokenization}.

\bibitem[{Tjong Kim~Sang and De~Meulder(2003)}]{tjong-kim-sang-de-meulder-2003-introduction}
Erik~F. Tjong Kim~Sang and Fien De~Meulder. 2003.
\newblock \href {https://aclanthology.org/W03-0419} {Introduction to the {C}o{NLL}-2003 shared task: Language-independent named entity recognition}.
\newblock In \emph{Proceedings of the Seventh Conference on Natural Language Learning at {HLT}-{NAACL} 2003}, pages 142--147.

\bibitem[{Vaswani et~al.(2017{\natexlab{a}})Vaswani, Shazeer, Parmar, Uszkoreit, Jones, Gomez, Kaiser, and Polosukhin}]{https://doi.org/10.48550/arxiv.1706.03762}
Ashish Vaswani, Noam Shazeer, Niki Parmar, Jakob Uszkoreit, Llion Jones, Aidan~N. Gomez, Lukasz Kaiser, and Illia Polosukhin. 2017{\natexlab{a}}.
\newblock \href {https://doi.org/10.48550/ARXIV.1706.03762} {Attention is all you need}.

\bibitem[{Vaswani et~al.(2017{\natexlab{b}})Vaswani, Shazeer, Parmar, Uszkoreit, Jones, Gomez, Kaiser, and Polosukhin}]{vaswani2017attention}
Ashish Vaswani, Noam Shazeer, Niki Parmar, Jakob Uszkoreit, Llion Jones, Aidan~N. Gomez, Lukasz Kaiser, and Illia Polosukhin. 2017{\natexlab{b}}.
\newblock \href {http://arxiv.org/abs/1706.03762} {Attention is all you need}.

\bibitem[{Wang et~al.(2021)Wang, Jiang, Bach, Wang, Huang, Huang, and Tu}]{wang-etal-2021-automated}
Xinyu Wang, Yong Jiang, Nguyen Bach, Tao Wang, Zhongqiang Huang, Fei Huang, and Kewei Tu. 2021.
\newblock \href {https://doi.org/10.18653/v1/2021.acl-long.206} {Automated concatenation of embeddings for structured prediction}.
\newblock In \emph{Proceedings of the 59th Annual Meeting of the Association for Computational Linguistics and the 11th International Joint Conference on Natural Language Processing (Volume 1: Long Papers)}, pages 2643--2660, Online. Association for Computational Linguistics.

\bibitem[{Wei et~al.(2023)Wei, Huang, Yu, and Liu}]{wei-etal-2023-ptcspell}
Xiao Wei, Jianbao Huang, Hang Yu, and Qian Liu. 2023.
\newblock \href {https://doi.org/10.18653/v1/2023.findings-acl.394} {{PTCS}pell: Pre-trained corrector based on character shape and {P}inyin for {C}hinese spelling correction}.
\newblock In \emph{Findings of the Association for Computational Linguistics: ACL 2023}, pages 6330--6343, Toronto, Canada. Association for Computational Linguistics.

\bibitem[{Wu et~al.(2016)Wu, Schuster, Chen, Le, Norouzi, Macherey, Krikun, Cao, Gao, Macherey, Klingner, Shah, Johnson, Liu, Łukasz Kaiser, Gouws, Kato, Kudo, Kazawa, Stevens, Kurian, Patil, Wang, Young, Smith, Riesa, Rudnick, Vinyals, Corrado, Hughes, and Dean}]{wu2016googles}
Yonghui Wu, Mike Schuster, Zhifeng Chen, Quoc~V. Le, Mohammad Norouzi, Wolfgang Macherey, Maxim Krikun, Yuan Cao, Qin Gao, Klaus Macherey, Jeff Klingner, Apurva Shah, Melvin Johnson, Xiaobing Liu, Łukasz Kaiser, Stephan Gouws, Yoshikiyo Kato, Taku Kudo, Hideto Kazawa, Keith Stevens, George Kurian, Nishant Patil, Wei Wang, Cliff Young, Jason Smith, Jason Riesa, Alex Rudnick, Oriol Vinyals, Greg Corrado, Macduff Hughes, and Jeffrey Dean. 2016.
\newblock \href {http://arxiv.org/abs/1609.08144} {Google's neural machine translation system: Bridging the gap between human and machine translation}.

\bibitem[{Xue et~al.(2022)Xue, Barua, Constant, Al-Rfou, Narang, Kale, Roberts, and Raffel}]{xue-etal-2022-byt5}
Linting Xue, Aditya Barua, Noah Constant, Rami Al-Rfou, Sharan Narang, Mihir Kale, Adam Roberts, and Colin Raffel. 2022.
\newblock \href {https://doi.org/10.1162/tacl_a_00461} {{B}y{T}5: Towards a token-free future with pre-trained byte-to-byte models}.
\newblock \emph{Transactions of the Association for Computational Linguistics}, 10:291--306.

\bibitem[{Zhu et~al.(2015)Zhu, Kiros, Zemel, Salakhutdinov, Urtasun, Torralba, and Fidler}]{zhu2015aligning}
Yukun Zhu, Ryan Kiros, Richard Zemel, Ruslan Salakhutdinov, Raquel Urtasun, Antonio Torralba, and Sanja Fidler. 2015.
\newblock \href {http://arxiv.org/abs/1506.06724} {Aligning books and movies: Towards story-like visual explanations by watching movies and reading books}.

\bibitem[{Zhuang et~al.(2021)Zhuang, Wayne, Ya, and Jun}]{zhuang-etal-2021-robustly}
Liu Zhuang, Lin Wayne, Shi Ya, and Zhao Jun. 2021.
\newblock \href {https://aclanthology.org/2021.ccl-1.108} {A robustly optimized {BERT} pre-training approach with post-training}.
\newblock In \emph{Proceedings of the 20th Chinese National Conference on Computational Linguistics}, pages 1218--1227, Huhhot, China. Chinese Information Processing Society of China.

\end{thebibliography}
\bibliographystyle{acl_natbib}

\clearpage

\appendix

\section{Positional Embeddings}
\label{app:positional-embeddings}
Since we are co-attending subword and character embeddings, it appears beneficial to re-introduce positional information in the co-attention modules. We do this by adding positional embeddings (PEs) to the character and subword embeddings output by the backbone encoder before they are passed to the co-attention modules. Since the character and subword sequence typically have different lengths, it is necessary to have strategies for translating between the character-level and subword-level PEs.  We consider three potential strategies: Using strategy A, each character inherits the PE of the subword it belongs to. Using strategy B, each subword inherits the PE of its first character. Using strategy C, the PE of each subword is the average of its character's PEs. 
Table \ref{tab:pos_embed_map} demonstrates the three strategies on a small example. We experiment with an entanglement model with 1 co-attention module and sinusoidal absolute PEs used by the original transformer \cite{vaswani2017attention}. 

\begin{table}[h]
    \begin{tabular}{l| c c c c c c c}
    \toprule
       \textbf{word position} & \textbf{A} &  \multicolumn{3}{l}{\textbf{dog}}  & \multicolumn{3}{l}{\textbf{sat}} \\ \midrule
        \rowcolor{yellow!20} 
         strategy A & 1 & \multicolumn{3}{l}{2} & \multicolumn{3}{l}{3} \\
        strategy B & 1 & \multicolumn{3}{l}{2} & \multicolumn{3}{l}{5} \\
        strategy C & 1 & \multicolumn{3}{l}{(2+3+4)/3} & \multicolumn{3}{l}{(5+6+7)/3} \\
        \midrule\midrule
        \textbf{char position} & \textbf{A} &  \textbf{d} & \textbf{o} & \textbf{g} & \textbf{s} & \textbf{a} & \textbf{t} \\ \midrule
        strategy A & 1 & 2 & 2 & 2 & 3 & 3 & 3 \\
        \rowcolor{yellow!20} 
        strategy B & 1 & 2 & 3 & 4 & 5 & 6 & 7 \\
        \rowcolor{yellow!20} 
        strategy C & 1 & 2 & 3 & 4 & 5 & 6 & 7 \\
        \bottomrule
    \end{tabular}
    \caption{Strategies for mappings the PEs. The un-highlighted PEs are derived from the highlighted PEs by the rule specified. }
    \label{tab:pos_embed_map}
\end{table}

\section{Pretraining}
\label{app:pretraining}

Since the co-attention modules are essentially transformer blocks, our model could be pretrained. To investigate the effect of pretraining on our model, we pretrain the model on a subset of the combined corpus of WikiText-103 \cite{merity2016pointer} and Bookcorpus \cite{zhu2015aligning}. The model is trained on three types of objectives: subword-level masked language modeling (MLM) loss, character-level MLM loss, and a novel character-word matching loss that aims to align the representation space of the output character and subword embeddings, described below. Table \ref{tab:pretrain} displays the results of the pre-trained model on WNUT-17 using different amount of data for pretraining, and we see that the model seems to perform worse when more data is used in pretraining. 


\subsection{Character-word Matching}
\begin{figure}[h]
    \centering
    \includegraphics[width = 7cm]{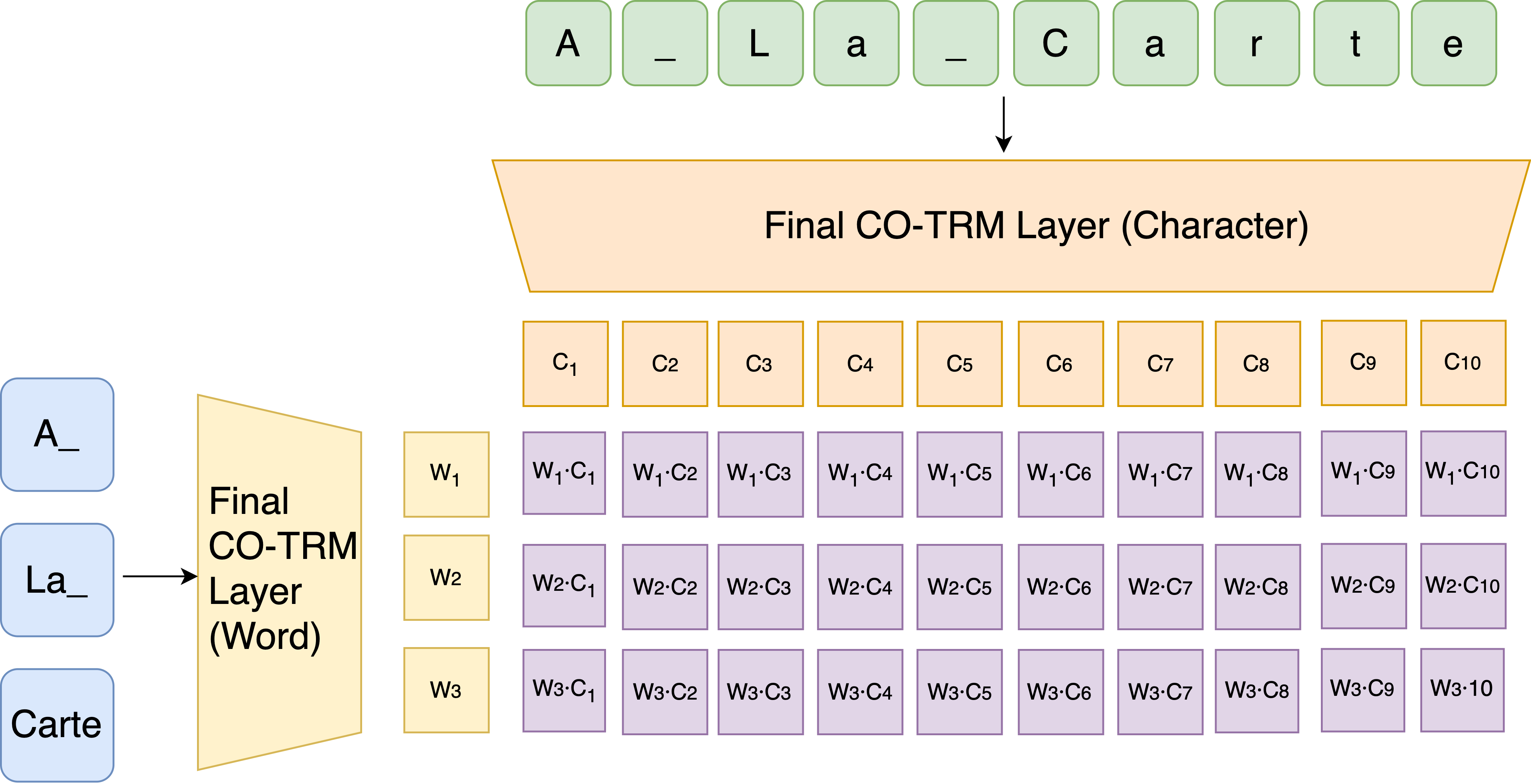}
    \caption{Character-word matching loss}
    \label{fig:matchingloss}
\end{figure}

In order to align the representation space of character and word embeddings, we propose a contrastive learning objective named character-subword matching, which is used during our pretraining step Figure \ref{fig:matchingloss} presents a visualization of the character-subword matching objective. For each character in $T^c$, we record a label for the subword that it belongs to. For example, consider the sentence \verb+A la carte+. Character \verb+A+ would be labeled 1 and  \verb+l,a+ would be labeled 2. We call the label sequence $L^c$

We compute the pairwise similarity (scaled dot product) between each subword-character pair, and we create a similarity matrix $S = H^s_* \cdot H^c_*$, where $S[i, j] = H^s_*[i] \cdot H^c_*[j] / a$, where $a$ is a trainable constant.  Therefore, we formulate the contrastive loss as follows:
$$
\mathcal{L}_c = \sum_{j=1}^m \text{CrossEntropyLoss}(S, \text{ref} = L^c)
$$

\subsection{Optimization}

To learn the parameters of our model, we optimize the model over three objectives. For MLM, we randomly masked out 15\% of the tokens in the subword and character sequence. Take a single piece of text $(\xv)$ for example. We respectively compute the subword-level and character-level MLM loss as follows:
\begin{align*}
&\mathcal{L}_{mlm}^{sub} (\xv)= -\sum_{t=1}^{n_i^s} \log(\xv^s_{t}|\xv^s_{\neq t}, \theta) \\
&\mathcal{L}_{mlm}^{char} (\xv)= - \sum_{t=1}^{n^c} \log(\xv^c_{ t}|\xv^c_{\neq t}, \theta) \\
\end{align*}
where $\xv^s_{t}, \xv^c_{t}$ are respectively subword and character tokens, and $n^s, n^c$ are respectively the number of subword tokens and character tokens. $\xv^s_{\neq t}$, $\xv^c_{\neq t}$ means the complete character/subword sequence without token $\xv^s_{ t}, \xv^c_{ t}$ and other masked-out tokens, and $\theta$ refers to the parameters in our model. 

Our model is then pretrained over the three objectives:
$$
\mathcal{L}(\xv) = \mathcal{L}_c(\xv) + \mathcal{L}_{mlm}^{sub} (\xv) + \mathcal{L}_{mlm}^{char} (\xv)
$$
We pretrain our model on a random subset of the combined corpus of WikiText-103 \cite{merity2016pointer} and Bookcorpus \cite{zhu2015aligning}. The model is trained for 8 hours on 4 Tesla V100 GPUs with 32 GB memory. The initial learning rate is 2e-5 and we used an Adam optimizer and a linear scheduler.

\begin{table}[t]
    \centering
    \begin{tabular}{l | c c c c c }
    \toprule
        \multicolumn{2}{l}{\textbf{Model}}  & \textbf{\% data} & \textbf{\# epoch} & \textbf{F1} \\ 
        \midrule
        \multicolumn{2}{l}{$\text{RoBERTa}_{\text{base}}$} & - & -  & 56.38\\
        \multicolumn{2}{l}{EM(\#C = 1)} & - & -  & \textbf{57.80}\\\midrule\midrule
        \textbf{Side} & \textbf{\#C} \\\midrule
        \wordside & 1 &  $\sim$ 0.18\% & 15  & {56.59 }\\
         & 1 &  $\sim$ 8\% & 1  & 51.71\\
         & 2 & $\sim$ 8\% & 1 & 52.25\\
         & 6 & $\sim$ 8\% & 1 & 53.14\\
         \bottomrule
    \end{tabular}
    \caption{F1 scores for pretrained entanglement model in WNUT-17. \% data refers to the \% of the corpus used for pretraining. EM(\#C = 1) refers to the un-pretrained entanglement model with 1 co-attention module.}
    \label{tab:pretrain}
\end{table}




\end{document}